\documentclass[letterpaper, 10 pt, conference]{ieeeconf}  

\IEEEoverridecommandlockouts                              

\usepackage{algorithm}
\usepackage{algorithmic}
\usepackage{amsmath}
\usepackage{booktabs}
\usepackage{multirow}
\usepackage{graphicx}
\usepackage{subcaption}
\usepackage{placeins}
\usepackage{color}

\usepackage{hyperref}

\title{\LARGE \bf
Class-Conditional Conformal Prediction for Signal Temporal Logic Inference with Many Classes
}

\author{%
Yixuan Wang$^{1}$, Danyang Li$^{2}$, Matthew Cleaveland$^{3}$, Roberto Tron$^{2}$, Mingyu Cai$^{1}$%
\thanks{$^{1}$Mingyu Cai and Yixuan Wang are with Mechanical Engineering, University of California, Riverside, CA 92521, USA.
{\tt\small \{mingyu.cai,ywang1457\}@ucr.edu}
$^{2}$Danyang Li and Roberto Tron are with Mechanical Engineering, Boston University.
{\tt\small \{danyangl,tron\}@bu.edu}
$^{3}$Matthew Cleaveland is with MIT Lincoln Laboratory, Lexington, MA 02421, USA.
{\tt\small \{matthew.cleaveland@ll\}@mit.edu}}%
}%

\begin{document}

\maketitle
\thispagestyle{empty}
\pagestyle{empty}

\begin{abstract}
Signal Temporal Logic (STL) inference provides interpretable formulas for time series classification. For multiclass classification, however, accurate predictions do not necessarily ensure that the inferred formulas are satisfied consistently with the true class. We introduce the positive satisfaction rate and codeword satisfaction rate to quantify satisfaction of required formulas and consistency with the complete class codeword, respectively. We develop a two-stage procedure that first learns discriminative formulas using a margin-based objective and then refines predicate thresholds toward a small positive robustness target while fixing the formula structure and temporal intervals. Satisfaction and ranking penalties discourage violations of required formulas and loss of class separation. The decoding loss is used as a nonconformity score and calibrated separately for each class, providing finite-sample class-conditional coverage under intra-class exchangeability. Experiments on naval, synthetic, and robot manipulation trajectories demonstrate discrepancies between classification accuracy and formula satisfaction. The results further reveal a data-dependent tradeoff between robustness tightness and satisfaction, while class-conditional calibration mitigates undercoverage that can be hidden by pooled calibration. The project page is available at \href{https://sites.google.com/ucr.edu/multiclasswithconditionalcp?usp=sharing}{Project Page}.

\end{abstract}

\section{INTRODUCTION}

Time series classification is widely used to identify system behaviors in autonomous and cyber-physical systems, including vessel motion, robot manipulation, and human activities~\cite{aasi2023time,liu2024timex++}. While neural classifiers can capture complex temporal patterns, they provide limited insight into the temporal properties that distinguish different classes. Temporal logic and neuro-symbolic approaches address this limitation by learning interpretable logical specifications, including Signal Temporal Logic (STL) formulas~\cite{maler2004monitoring, donze2010robust,li2024tlinet,liu2024interpretable,ferfoglia2025towards}. Most STL inference methods focus on binary classification, while multiclass STL models learn multiple formulas whose robustness values are used to distinguish among class specific specifications~\cite{li2024multi}.

Existing multiclass STL models use margin based objectives to separate class dependent robustness patterns~\cite{li2024multi}. Tight STL inference instead seeks formulas that remain satisfied while approaching the satisfaction boundary~\cite{jha2019telex}. These objectives concern different properties. Classification depends on the relative ordering of class decoding losses, whereas codeword satisfaction requires the robustness signs to agree with every entry of the true class codeword. Accuracy alone therefore cannot establish whether the learned formulas are satisfied according to the intended class codewords.
Another limitation concerns uncertainty in the resulting multiclass predictions. A point prediction does not indicate which other classes should remain in consideration for a trajectory. Conformal prediction constructs prediction sets with finite sample coverage guarantees under exchangeability~\cite{vovk2005algorithmic}. Recent work has also begun to study statistical reliability in temporal-logic learning, including finite-sample guarantees for learned temporal predicates and conformal calibration for STL inference~\cite{soroka2024learning,li2025conformal}. In multiclass classification, adaptive and rank based conformal scores seek smaller prediction sets while retaining marginal coverage~\cite{romano2020classification,huang2023conformal}. Marginal coverage, however, does not guarantee the target coverage separately for each class. Classwise and group conditional methods address this limitation, including settings with imbalanced classes or many labels~\cite{shi2024conformal,bairaktari2025kandinsky,gao2025bridging}. For multiclass STL models, the decoding loss provides a score for each candidate class that can be calibrated separately within that class. 
We address these two limitations through a two stage multiclass STL inference framework followed by class conditional conformal calibration. The main contributions are as follows:

\begin{itemize}
    \item We introduce two satisfaction rates for multiclass STL models, measuring satisfaction over required class properties and complete class specifications, respectively, to reveal satisfaction degradation not reflected by classification accuracy.
    \item We develop a two-stage learning procedure that separates discriminative formula learning from tightness refinement. The refinement stage adjusts predicate thresholds toward the satisfaction boundary using satisfaction and ranking penalties to balance tightness, satisfaction, and class separation, with a tradeoff weight.
    \item We combine the learned STL model with class-conditional conformal prediction, using the class specific loss as the nonconformity score and estimating a separate calibration threshold for each class, providing finite-sample class-conditional coverage under within-class exchangeability.
\end{itemize}

\noindent\textbf{Related Work.} 
Earlier work mines STL requirements from closed loop control models ~\cite{7084172}. Decision tree methods construct classifiers from parameterized STL primitives~\cite{bombara2016decision}, while enumerative methods search over candidate formula structures and synthesize their parameters~\cite{mohammadinejad2020interpretable}. STLCG represents STL robustness using computation graphs to support gradient based optimization
~\cite{leung2023backpropagation}. Neural approaches use differentiable STL modules to learn formula structures and parameters through gradient-based
optimization~\cite{li2024tlinet}. Related neural methods learn weighted graph based STL
formulas~\cite{9662049}. Multiclass STL classification has also been studied using weighted STL representations with neural learning and decision tree construction~\cite{yan2022neuro}.
Many STL classification methods focus on binary labels. A multiclass framework learns attribute formulas and uses their robustness values to distinguish among class specific specifications~\cite{li2024multi}. Its margin objective promotes correctly signed and separated robustness values. We build on this framework and evaluate both satisfaction of required formulas and consistency with the complete class specification. Information guided temporal logic inference also uses
prior knowledge to favor informative descriptions
~\cite{8815145}. In contrast, tight specification learning seeks formulas that characterize observed trajectories without unnecessary robustness slack. TeLEx learns parameters of a given STL template from positive examples using a smooth tightness metric~\cite{jha2019telex}. Unlike positive-only template fitting, our refinement starts from multiclass discriminative learning and incorporates satisfaction and ranking penalties.

Uncertainty aware STL inference represents uncertain
trajectories as intervals and optimizes worst case
robustness~\cite{baharisangari2021uncertainty}. Conformal prediction provides finite-sample marginal coverage under exchangeability~\cite{vovk2005algorithmic}. Adaptive methods construct compact prediction sets by accounting for differences in prediction difficulty~\cite{romano2020classification,huang2023conformal}, while conditional approaches target coverage within specified groups~\cite{gibbs2025conformal}. Conformal prediction has also been integrated with STL inference using robustness- and margin-based scores~\cite{li2025conformal}. In this work, we calibrate after discriminative learning and tightness refinement, using the multiclass classification loss as the nonconformity score and estimating a separate threshold for each class to obtain class-conditional coverage.

\section{Problem Formulation}
\label{sec:problem formulation}
\subsection{STL-Based Multiclass Classification}
Let $x:\mathbf{T}\rightarrow\mathbf{R}^{d}$ be a $d$-dimensional discrete-time signal defined on $\mathbf{T}=\{0,1,\ldots,T\}$. Signal Temporal Logic (STL) specifies temporal properties of real-valued signals. We consider formulas generated by
\[
\varphi
::=
\mu
\mid
\neg\varphi
\mid
\varphi_1\wedge\varphi_2
\mid
\varphi_1\vee\varphi_2
\mid
\mathbf{F}_{I}\varphi
\mid
\mathbf{G}_{I}\varphi,
\]
where $\mu$ is an atomic predicate and $I=[a,b]\subseteq\mathbf{Z}_{\geq 0}$ is a bounded
discrete-time interval. The operators $\mathbf{F}_{I}$ and $\mathbf{G}_{I}$ denote eventually and always, respectively. An atomic predicate is written as $\mu(x(t)):=h(x(t))\geq0$. In this work, $h(z)=\pm(z_j-\theta)$, so each predicate compares one signal component with a threshold. The relation $x,t\models\varphi$ denotes satisfaction of $\varphi$ by $x$ at time $t$. Let $t+I=\{t+\tau:\tau\in I\}$. We consider formulas whose temporal horizons lie within the available signal horizon and evaluate them at $t=0$, writing
$x\models\varphi$ for $x,0\models\varphi$.


STL also assigns a real-valued robustness score $\rho(x,\varphi,t)\in\mathbf{R}$ to a signal and a formula. For an atomic predicate, $\rho(x,\mu,t)=h(x(t))$, and the robustness of the Boolean and temporal operators is defined recursively as
\[
\rho(x,\neg\varphi,t)
=
-\rho(x,\varphi,t),
\]
\[
\rho(x,\varphi_1\wedge\varphi_2,t)
=
\min\{
\rho(x,\varphi_1,t),
\rho(x,\varphi_2,t)
\},
\]
\[
\rho(x,\varphi_1\vee\varphi_2,t)
=
\max\{
\rho(x,\varphi_1,t),
\rho(x,\varphi_2,t)
\},
\]
\[
\rho(x,\mathbf{F}_{I}\varphi,t)
=
\max_{t'\in(t+I)\cap\mathbf{T}}
\rho(x,\varphi,t'),
\]
\[
\rho(x,\mathbf{G}_{I}\varphi,t)
=
\min_{t'\in(t+I)\cap\mathbf{T}}
\rho(x,\varphi,t').
\]
For formulas evaluated at the initial time, we write $\rho(x,\varphi):=\rho(x,\varphi,0)$. The sign of the robustness is consistent with the Boolean semantics. If $\rho(x,\varphi)>0$, then $x\models\varphi$, whereas if $\rho(x,\varphi)<0$, then $x\not\models\varphi$. Its magnitude quantifies the degree of satisfaction or violation under the chosen robustness semantics.

Multiclass classification can be formulated by learning $K$ real-valued functions whose outputs jointly characterize $C$ classes~\cite{dietterich1994solving,allwein2000reducing}. Let $M\in{-1,+1}^{C\times K}$ denote the desired class-property patterns, where $M_{c,k}$ specifies the desired sign of the $k$-th output for class $c$. Given the output vector
$f(x)=\big(f_1(x),\ldots,f_K(x)\big),$
classification can be performed by measuring its disagreement with each class pattern. A loss-based rule selects
$\hat y(x)=\arg\min_{c} d(x,c),$
where
$d(x,c)=\sum_{k=1}^{K}\ell\big(M_{c,k}f_k(x)\big)$
and $\ell(\cdot)$ penalizes disagreement with the desired sign~\cite{allwein2000reducing}. In multiclass STL inference~\cite{li2024multi}, each $f_k(x)$ corresponds to the robustness of an STL formula $\varphi_k$, so its sign indicates whether the associated temporal property is satisfied or violated. Thus, multiple STL formulas jointly characterize each class, while their robustness values provide both class predictions and quantitative measures of property satisfaction.

\subsection{Conformal Prediction}
Let $(X,Y)\in\mathcal{X}\times\mathcal{Y}$ denote an input-label pair. Given a
calibration set and a nonconformity score $s(x,c)$, where larger values indicate
lower compatibility with class $c$, split conformal prediction constructs a
prediction set
\begin{equation}
    \mathcal{C}_{\alpha}(x) = \{c : s(x,c) \leq \widehat{q}\},
\end{equation}
For calibration scores $S_i=s(X_i,Y_i)$, $i=1,\ldots,n$, set $\widehat q=S_{(k)}$, where $k=\lceil(n+1)(1-\alpha)\rceil$ and $S_{(k)}$ is the
$k$-th smallest score. Set $\widehat q=+\infty$ if $k>n$. Under exchangeability, the prediction set satisfies the marginal coverage guarantee
\begin{equation}
    \mathbf{P}\!\left(Y \in \mathcal{C}_{\alpha}(X)\right) \geq 1 - \alpha.
\end{equation}
However, marginal coverage does not guarantee the same coverage across different
classes.

Let $\mathcal I_c=\{i:Y_i=c\}$ and $n_c=|\mathcal I_c|$. For the classwise scores $S_i^c=s(X_i,c)$, $i\in\mathcal I_c$, set $\widehat q_c=S^c_{(k_c)}$, where $k_c=\lceil(n_c+1)(1-\alpha)\rceil$. Set $\widehat q_c=+\infty$ if $k_c>n_c$. These class specific thresholds give prediction sets
\begin{equation}
    \mathcal{C}^{\mathrm{cc}}_{\alpha}(x) = \{c : s(x,c) \leq \widehat{q}_c\}.
\end{equation}
Under exchangeability within each class, this provides the class-conditional
guarantee
\begin{equation}
    \mathbf{P}\!\left(Y \in \mathcal{C}^{\mathrm{cc}}_{\alpha}(X) \mid Y = c\right)
    \geq 1 - \alpha.
\end{equation}
With finite calibration data, class specific thresholds become less precise when the number of calibration samples per class is small, and the realized coverage varies more across calibration sets, since the coverage is a random variable over the calibration data.

\subsection{Problem Setup}

\noindent\textbf{C.1. Data splits.}
Let $\mathcal{D}_{\mathrm{tr}}$, $\mathcal{D}_{\mathrm{val}}$, $\mathcal{D}_{\mathrm{cal}}$,
and $\mathcal{D}_{\mathrm{test}}$ be four disjoint sets of labeled signals
$(x_i, y_i)$ with $y_i \in \mathcal{Y} = \{1,\ldots,C\}$, used respectively for
formula learning, model selection, conformal calibration, and final evaluation.

\noindent\textbf{C.2. STL classifier.}
The classifier uses $K$ STL formulas $\{\varphi_1,\ldots,\varphi_K\}$ and a fixed
code matrix $M \in \{+1,-1\}^{C \times K}$, where $M_{c,k} = +1$ requires class
$c$ signals to satisfy $\varphi_k$ and $M_{c,k} = -1$ requires violation. Let
$\widetilde{R}_{\theta,k}(x)$ denote the differentiable robustness approximation of
$\varphi_k$, parameterized by $\theta$. The predicted class minimizes the decoding
loss
\begin{equation}
    d_{\theta}(x,c) = \sum_{k=1}^{K}
    \left[-M_{c,k}\,\widetilde{R}_{\theta,k}(x)\right]_{+},
    \label{eq:decode}
\end{equation}
which penalizes disagreement between model outputs and required code entries.

\noindent\textbf{C.3. Satisfaction metrics.}
Let $\mathcal{P}_{\mathcal{D}} = \{(i,k) \mid M_{y_i,k} = +1\}$ index the
formula--signal pairs required to be satisfied. The positive satisfaction rate
(PSR) and codeword satisfaction rate (CSR) measure how well model outputs agree
with the code matrix:
\begin{equation}
\begin{aligned}
    \mathrm{PSR}(\mathcal{D})
    &= \frac{1}{|\mathcal{P}_{\mathcal{D}}|}
       \sum_{(i,k)\in\mathcal{P}_{\mathcal{D}}}
       \mathbf{1}\!\left\{\widetilde{R}_{\theta,k}(x_i) > 0\right\}, \\[4pt]
    \mathrm{CSR}(\mathcal{D})
    &= \frac{1}{n_{\mathcal{D}}}
       \sum_{i=1}^{n_{\mathcal{D}}}
       \mathbf{1}\!\left\{
         \min_{k} M_{y_i,k}\,\widetilde{R}_{\theta,k}(x_i) > 0
       \right\}.
\end{aligned}
\label{eq:metrics}
\end{equation}
PSR captures per-formula satisfaction among required entries; CSR requires full codeword agreement including required violations. We also report the mean robustness over positive code entries,
\[
\bar R_{\mathrm{pos}}(\mathcal D)
=
\frac{1}{|\mathcal P_{\mathcal D}|}
\sum_{(i,k)\in\mathcal P_{\mathcal D}}
\widetilde R_{\theta,k}(x_i).
\]
This mean includes violated entries, so a decrease alone does not establish tighter satisfaction and must be interpreted alongside PSR and CSR.

\noindent\textbf{C.4. Motivation and challenges.}
STL formulas describe temporal behavior through explicit logical conditions, but classification accuracy alone does not show whether signals satisfy the formulas required by their classes. Discriminative
learning also does not explicitly favor small positive robustness. Refining predicate thresholds can reduce robustness slack, but may cause some signals to violate required formulas or change their
predicted classes. These effects motivate evaluating satisfaction alongside accuracy during refinement. A point prediction provides no coverage guarantee on its own, motivating conformal prediction sets with coverage calibrated separately for each class.

\noindent\textbf{C.5. Objectives.}
The overall goal is threefold:
\begin{enumerate}
    \item \textbf{Formula learning.} Train $\theta$ on $\mathcal{D}_{\mathrm{tr}}$
    to minimize decoding loss and achieve accurate multiclass prediction.

    \textbf{Thresho ld adjustment.} Refine predicate thresholds toward a small positive robustness target while limiting losses in satisfaction and class separation.

    \item \textbf{Conformal prediction sets.} Using $\mathcal{D}_{\mathrm{cal}}$,
    construct class-conditional prediction sets
    $\mathcal{C}^{\mathrm{cc}}_{\alpha}(x)$ satisfying
    \begin{equation}
        \mathbf{P}\!\left(Y \in \mathcal{C}^{\mathrm{cc}}_{\alpha}(X)
        \mid Y = c\right) \geq 1 - \alpha, \qquad \forall\, c \in \mathcal{Y},
        \label{eq:coverage}
    \end{equation}
    for a prescribed $\alpha \in (0,1)$. This coverage guarantee concerns
    inclusion of the true label rather than formula satisfaction, and relies on
    exchangeability of calibration and test examples within each class.
\end{enumerate}

\begin{figure}[!ht]
\centering
\includegraphics[width=0.88\columnwidth]{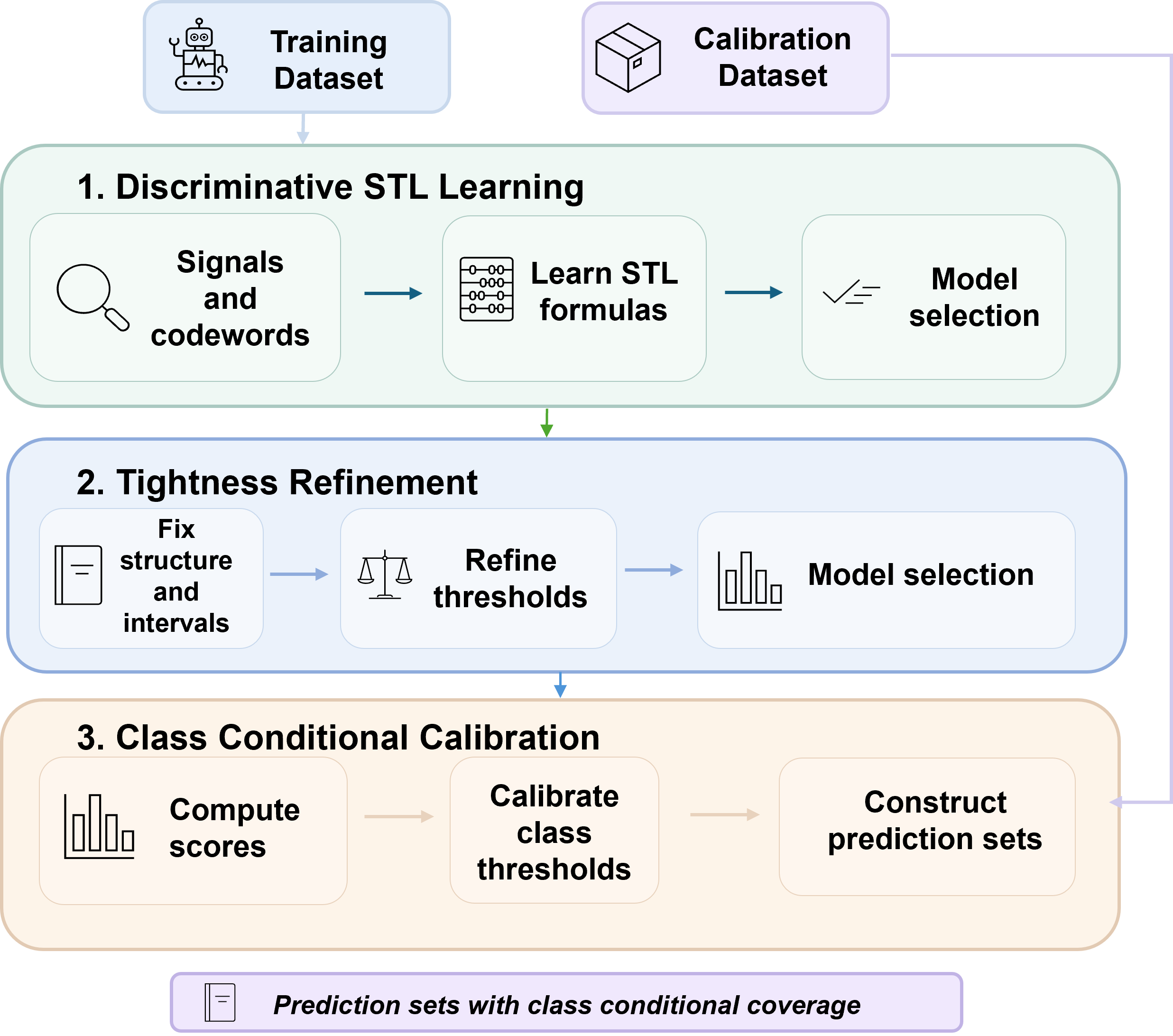}
\caption{Overview of the proposed framework combining discriminative STL learning, tightness refinement, and class conditional conformal prediction.}
\label{fig:framework}
\end{figure}

\vspace{-1.2em}

\section{Methodology}

The method proceeds in three steps. Discriminative learning trains formulas whose signs follow the codewords and whose decoding losses rank the true class first. Refinement then keeps the formula structure and moves only the thresholds, pulling satisfied required formulas toward a small positive robustness, while satisfaction, ranking, and anchor penalties limit the loss of satisfaction, the loss of class separation, and threshold drift. The selected model is finally calibrated class by class, with its decoding loss as the nonconformity score, as shown in Fig~\ref{fig:framework}.

\subsection{Discriminative Multiclass STL Learning}

The formulas follow the architecture of~\cite{li2024tlinet}, with each formula represented
as a disjunction of conjunctions of eventually or always subformulas over threshold predicates. Logical and temporal operations use differentiable approximations, and discrete logical connections are trained using a straight through estimator.

We adopt the margin construction of~\cite{li2024multi}. For each formula, its margin is the smallest nonnegative signed output in a minibatch $m_k=\min_{1\leq i\leq N} [M_{y_i,k}\widetilde{R}_{\theta,k}(x_i)]_{+}$, and $\overline{m}=K^{-1}\sum_{k=1}^{K}m_k$. For a minibatch, let $\mathcal P=\{(i,k):M_{y_i,k}=+1\}$ and
$\mathcal N=\{(i,k):M_{y_i,k}=-1\}$. The codewise losses are
\[
\begin{aligned}
\mathcal L_{\mathrm{pos}}
&=
\frac{1}{|\mathcal P|}
\sum_{(i,k)\in\mathcal P}
[m_k-\widetilde R_{\theta,k}(x_i)]_+,
\\
\mathcal L_{\mathrm{neg}}
&=
\frac{1}{|\mathcal N|}
\sum_{(i,k)\in\mathcal N}
[m_k+\widetilde R_{\theta,k}(x_i)]_+.
\end{aligned}
\] 
If some output of formula $k$ has the wrong sign, $m_k=0$ and the hinges reduce to penalties for sign disagreement. Otherwise every hinge and every such penalty is zero. In both cases the hinges equal sign penalties, and the margin enters the objective only through the reward $-\delta\overline{m}$, whose gradients propagate through $\overline{m}$.
The weight $\lambda_{\mathrm{neg}}$ controls the contribution of required violations relative to required satisfactions. These violations are important when signals from different classes satisfy the same formulas, since positive satisfaction alone cannot distinguish those classes.

The codewise losses act on each formula separately, whereas prediction depends on differences between class decoding losses. We therefore also penalize insufficient separation between these losses. For signal $x_i$, define
\[
\Delta_i(\theta)
=
\min_{c\in\mathcal{Y}\setminus\{y_i\}}
d_{\theta}(x_i,c)
-
d_{\theta}(x_i,y_i),
\]
which is positive when the true class has a smaller decoding loss than every competing class. The ranking penalty
\begin{equation}
\mathcal{L}_{\mathrm{rank}}(\theta)
=
\frac{1}{N}
\sum_{i=1}^{N}
[\mu-\Delta_i(\theta)]_{+}^{2},
\qquad \mu>0,
\label{eq:rank}
\end{equation}
encourages the true class to lead its closest competitor by at least $\mu$. The discriminative objective is
\begin{equation}
\begin{aligned}
\mathcal{L}_{\mathrm{disc}}
={}&
\mathcal{L}_{\mathrm{pos}}
+
\lambda_{\mathrm{neg}}\mathcal{L}_{\mathrm{neg}}
\\
&-
\delta\overline{m}
+
\lambda_{\mathrm{rank}}
\mathcal{L}_{\mathrm{rank}}(\theta),
\end{aligned}
\label{eq:disc}
\end{equation}
with $\delta>0$ and nonnegative remaining weights. The codewise hinges and the margin reward follow~\cite{li2024multi}. This work
separates and normalizes the positive and negative contributions, weights the latter by $\lambda_{\mathrm{neg}}$, and adds the ranking penalty. The discriminative model is selected by validation accuracy.

\subsection{Tightness Refinement}

A formula satisfied with large robustness may impose conditions that are less restrictive than the observed signals permit. Tightness seeks conditions that the signals satisfy
closer to the boundary, following the motivation of TeLEx~\cite{jha2019telex}.
The discriminative objective does not explicitly favor this proximity. Refinement fixes the selected logical structure and temporal intervals and adjusts only the predicate thresholds $b\in\mathbf{R}^{n_b}$. We write $\widetilde{R}_{b,k}$ and $d_b$ for the model outputs and decoding loss with all other parameters fixed.

The tightening term applies the TeLEx peak function
\[
T_{\beta}(r)
=
\frac{1}{r+e^{-\beta r}}-e^{-r}
\]
to the positive outputs of formulas required by the true class. For a minibatch, let $\mathcal{P}=\{(i,k)\mid M_{y_i,k}=+1\}$.
Then
\begin{equation}
\mathcal{L}_{\mathrm{tight}}
=
-\frac{1}{|\mathcal{P}|}
\sum_{(i,k)\in\mathcal{P}}
T_{\beta}
\left(
[\widetilde{R}_{b,k}(x_i)]_{+}
\right).
\label{eq:tight}
\end{equation}
For the value $\beta=20$ used here, the peak occurs at robustness approximately $0.15$ in normalized units. As a function of an individual positive output, this loss favors movement toward that target, lowering outputs above the peak and raising those below it.
It therefore favors a small positive robustness rather than minimizing every positive output toward zero. Negative outputs receive no direct gradient from this term.

Each predicate threshold is shared across signals. An adjustment that reduces large positive outputs can also reduce outputs of signals closer to the boundary, causing some required satisfactions to become violations. A separate satisfaction penalty discourages this effect,
\begin{equation}
\mathcal{L}_{\mathrm{sign}}
=
\frac{1}{|\mathcal{P}|}
\sum_{(i,k)\in\mathcal{P}}
[\gamma-\widetilde{R}_{b,k}(x_i)]_{+}^{2},
\qquad \gamma>0.
\label{eq:satpenalty}
\end{equation}
It penalizes existing violations and positive outputs below the satisfaction margin $\gamma$.
The weight of this term determines how strongly refinement favors satisfaction relative to the tightening target.

The ranking penalty in~\eqref{eq:rank} is recomputed using $d_b$, the decoding loss of Eq.~\eqref{eq:decode} with only the thresholds $b$ free. It discourages the true class from losing its advantage over competing classes as the thresholds change. It does not require the original decoding gaps or the ordering among competing classes to remain unchanged. Threshold movement from the selected discriminative model is additionally controlled by
$\mathcal{L}_{\mathrm{anchor}}
=\|b-b^{\mathrm{disc}}\|_2^2/n_b$.
The refinement objective is
\begin{equation}
\begin{aligned}
\mathcal{L}_{\mathrm{ref}}
={}&
\eta_{\mathrm{tight}}\mathcal{L}_{\mathrm{tight}}
+
\eta_{\mathrm{sign}}\mathcal{L}_{\mathrm{sign}}
\\
&+
\eta_{\mathrm{rank}}\mathcal{L}_{\mathrm{rank}}(b)
+
\eta_{\mathrm{anchor}}\mathcal{L}_{\mathrm{anchor}},
\end{aligned}
\label{eq:ref}
\end{equation}
with nonnegative weights.

These terms balance proximity to the positive robustness target, satisfaction, and class separation. They encourage these properties without imposing hard preservation constraints. Since tightening acts on formula outputs, it also does not require every constituent predicate to approach its own boundary. Refined candidates are selected on $\mathcal{D}_{\mathrm{val}}$. Candidates must stay within a validation accuracy tolerance of the
discriminative model, and selection then uses satisfaction or robustness over satisfied positive entries, as specified in the experimental settings. If no candidate meets the criteria, the discriminative model is retained. The selected model is fixed before calibration.

\subsection{Class Conditional Conformal Calibration}

Let $\theta^\star$ denote the selected model parameters. We use $s(x,c)=d_{\theta^\star}(x,c)$ as the nonconformity score and estimate a threshold $\widehat{q}_c$ from calibration examples of class $c$ using the procedure defined in Sec~\ref{sec:problem formulation}. The resulting prediction set is
\[
\mathcal{C}_{\alpha}^{\mathrm{cc}}(x)
=
\left\{
c\in\mathcal{Y}
\mid
d_{\theta^\star}(x,c)\leq\widehat{q}_c
\right\}.
\]

The score is zero exactly when $M_{c,k}\widetilde{R}_{\theta^\star,k}(x)\geq0$
for every $k$. Away from zero valued outputs, this condition agrees with the strict class satisfaction measured by CSR. Thus, the calibration score directly reflects disagreement with the class satisfaction pattern.

This connection also explains when a class threshold can be zero. Let $z_c$ be the number of zero scores among the $n_c$ calibration examples of class $c$. Then $\widehat{q}_c=0$ precisely when $z_c\geq\lceil(n_c+1)(1-\alpha)\rceil$. For such a class, inclusion in the prediction set requires nonnegative signed outputs for its entire satisfaction pattern. When all class thresholds are zero, a signal that matches no pattern in this sense receives an empty set. Under exchangeability within each class, calibration provides the coverage guarantee stated in Sec~\ref{sec:problem formulation}. The guarantee applies to inclusion of the true class under the fixed model score.

\begin{figure*}[t]
\centering

\begin{subfigure}{0.25\textwidth}
  \centering
  \includegraphics[width=\linewidth,height=0.145\textheight]{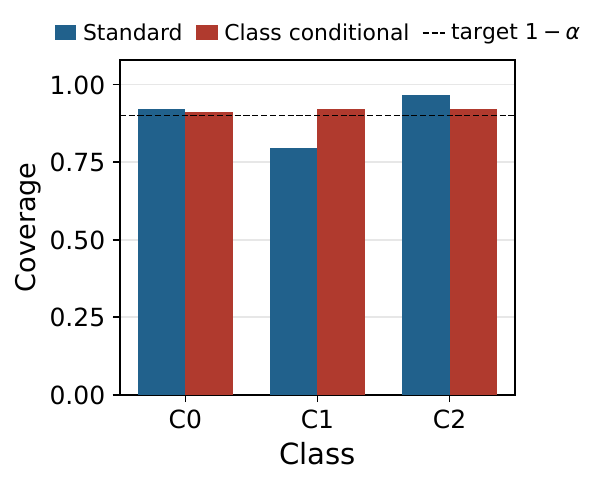}
\end{subfigure}
\hspace{0.02\textwidth}
\begin{subfigure}{0.25\textwidth}
  \centering
  \includegraphics[width=\linewidth,height=0.145\textheight]{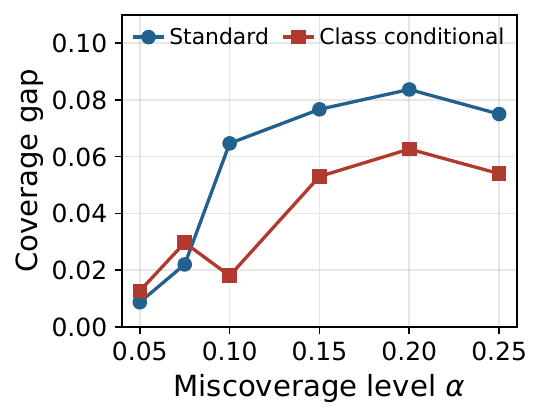}
\end{subfigure}
\hspace{0.02\textwidth}
\begin{subfigure}{0.25\textwidth}
  \centering
  \includegraphics[width=\linewidth,height=0.145\textheight]{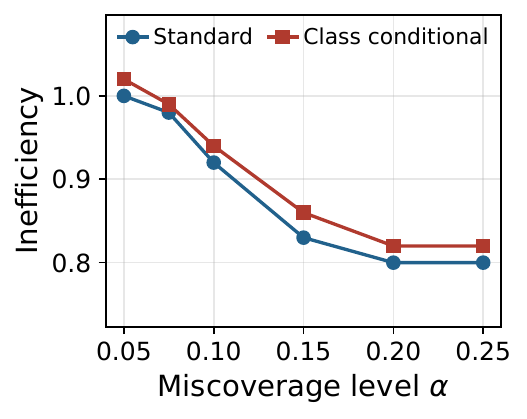}
\end{subfigure}

\vspace{-0.35em}

\begin{subfigure}{0.25\textwidth}
  \centering
  \includegraphics[
    width=1.05\linewidth,
    height=0.135\textheight
  ]{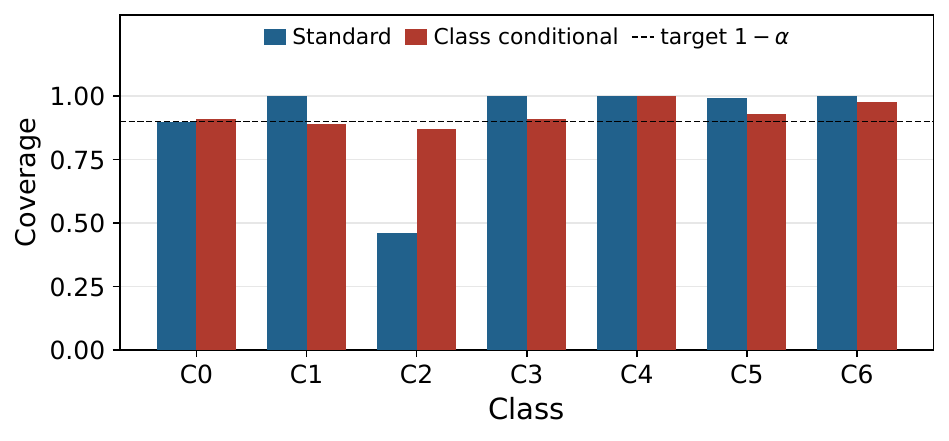}
\end{subfigure}
\hspace{0.02\textwidth}
\begin{subfigure}{0.25\textwidth}
  \centering
  \includegraphics[width=\linewidth,height=0.135\textheight]{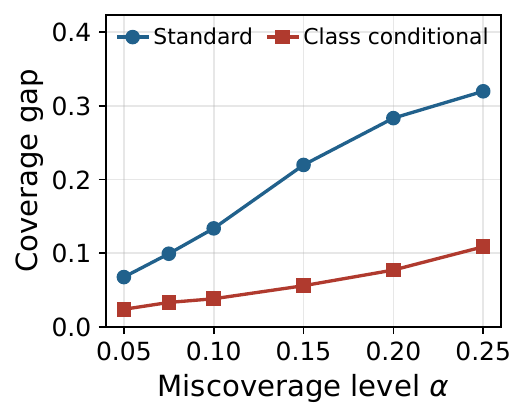}
\end{subfigure}
\hspace{0.02\textwidth}
\begin{subfigure}{0.25\textwidth}
  \centering
  \includegraphics[width=\linewidth,height=0.135\textheight]{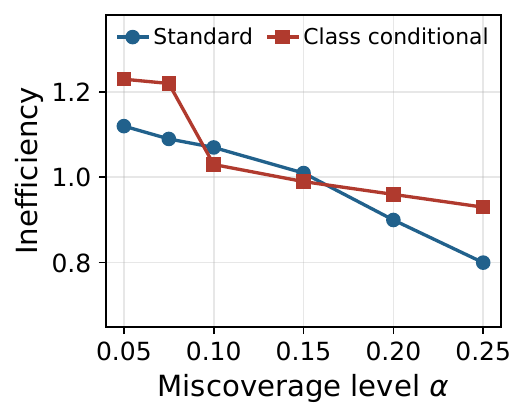}
\end{subfigure}

\vspace{-0.35em}

\begin{subfigure}{0.25\textwidth}
  \centering
  \includegraphics[
    width=1.05\linewidth,
    height=0.135\textheight
  ]{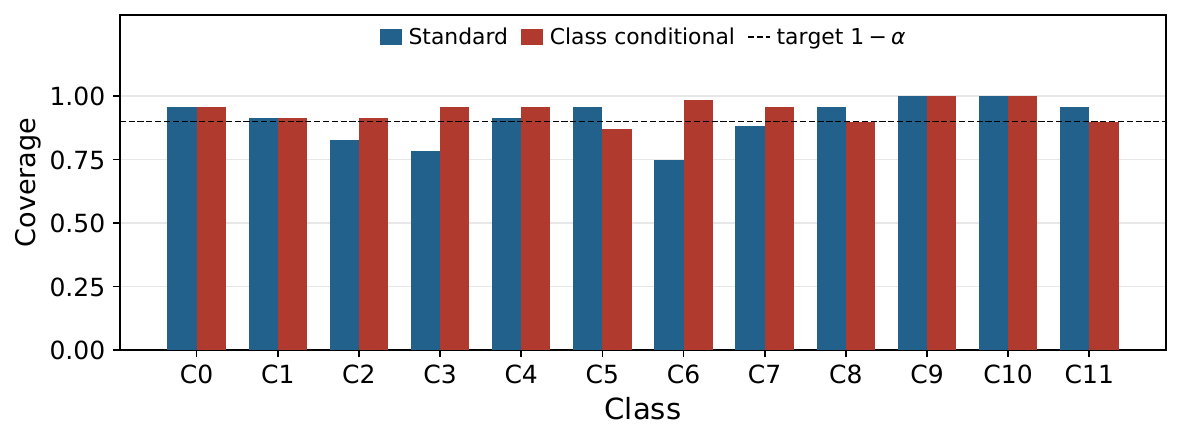}
  \caption{Classwise coverage at $\alpha=0.10$.}
\end{subfigure}
\hspace{0.02\textwidth}
\begin{subfigure}{0.25\textwidth}
  \centering
  \includegraphics[width=\linewidth,height=0.135\textheight]{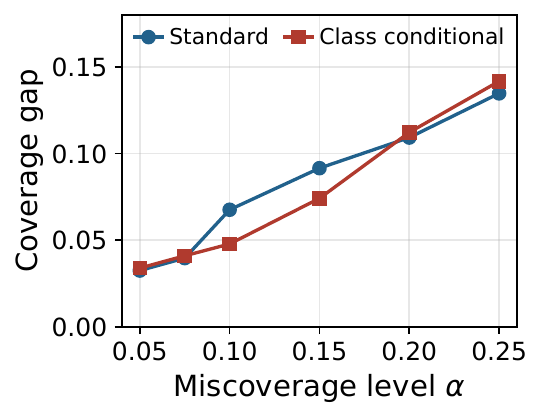}
  \caption{Coverage gap.}
\end{subfigure}
\hspace{0.02\textwidth}
\begin{subfigure}{0.25\textwidth}
  \centering
  \includegraphics[width=\linewidth,height=0.135\textheight]{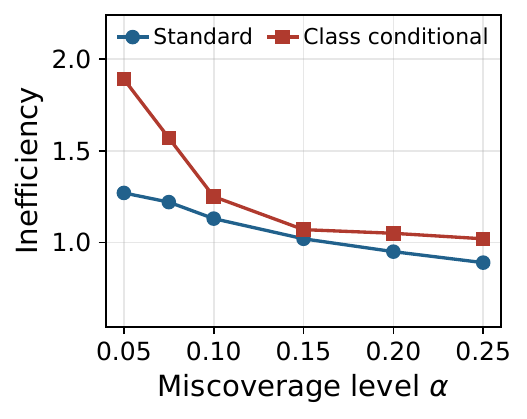}
  \caption{Inefficiency.}
\end{subfigure}

\vspace{-0.6em}

\caption{Conformal prediction with the refined models, standard against class conditional calibration. Rows: naval, synthetic, and CALVIN datasets. Columns: classwise coverage at $\alpha=0.10$, coverage gap, and inefficiency across miscoverage levels. Coverage gap is the average absolute difference between the classwise
coverage and $1-\alpha$, and inefficiency is the average prediction set size.}
\label{fig:cp-all}
\end{figure*}
\section{CASE STUDY}
We evaluate on three multiclass trajectory datasets, which we refer to as the naval, synthetic, and CALVIN datasets. The naval dataset contains $2000$ planar vessel trajectories in three classes, split $1200/300/249/251$ into training, validation, calibration, and test sets. The synthetic dataset contains $17500$ trajectories in seven classes, each generated from a conjunction of box shaped spatio temporal requirements, split $10500/2625/2184/2191$. The CALVIN dataset is derived from the CALVIN robot manipulation benchmark~\cite{mees2022calvin}, with $3620$ end effector trajectories of twelve tabletop manipulation tasks, split $2165/543/456/456$. CALVIN segments that share recorded frames are grouped before splitting, so no two splits share any frame. All experiments use one hot codes with $K = C$, with $\lambda_{\mathrm{neg}} = 1$ on the naval and CALVIN datasets and $\lambda_{\mathrm{neg}} = 0.1$ on the synthetic dataset. Accuracy, PSR, CSR, and $\bar{R}_{\mathrm{pos}}$ are reported on the test split, and the w/o margin ablation of the methodology accompanies the naval and synthetic experiments. Stage 1 retains the checkpoint with the highest validation accuracy. Stage 2 checkpoints must stay within 0.08 of the Stage 1 validation accuracy; on the naval and CALVIN datasets they must also reach PSR 0.40, and the checkpoint with the lowest mean robustness over the satisfied positive entries is retained, while on the synthetic dataset the highest PSR is retained. If no checkpoint passes the gates, the discriminative model stands.

The margin reward weight is $\delta = 0.1$ throughout. The satisfaction weight is swept over $\eta_{\mathrm{sign}} \in \{5, 10, 20\}$ on the naval dataset, with $\eta_{\mathrm{sign}} = 20$ as the reported operating point, and the satisfaction margin is $\gamma = 0.01$. Signals are normalized per channel to $[0,1]$ with constants computed on the training split only, and all reported thresholds are mapped back to the original units, namely the raw coordinate frame for the naval dataset, the generator frame for the synthetic dataset, and centimeters for CALVIN. CALVIN segments are resampled to 64 steps by linear interpolation. In all three experiments the network and exact robustness signs agree on every code entry, so the reported satisfaction rates coincide under both semantics. Accuracies under the two semantics differ because decoding compares loss magnitudes.


\begin{table}[t]
\centering
\caption{Accuracy, satisfaction rates, and mean positive robustness
$\bar{R}_{\mathrm{pos}}$ on the test split. Refinement reduces
$\bar{R}_{\mathrm{pos}}$ toward zero. The w/o margin rows remove the
margin term of Eq.~\eqref{eq:disc}, reducing the codewise losses to
sign losses.}
\label{tab:main}

\setlength{\tabcolsep}{3.5pt}
\scriptsize
\resizebox{\columnwidth}{!}{%
\begin{tabular}{llccccc}
\toprule
Dataset & Model & Acc$_{\mathrm{val}}$ & Acc$_{\mathrm{test}}$ &
PSR$_{\mathrm{test}}$ & CSR$_{\mathrm{test}}$ & $\bar{R}_{\mathrm{pos}}$ \\
\midrule
\multirow{3}{*}{Naval}
 & Discriminative & 0.977 & 0.980 & 0.924 & 0.872 & 0.24 \\
 & Refined        & 0.960 & 0.960 & 0.873 & 0.797 & 0.13 \\
 & w/o margin     & 0.997 & 1.000 & 0.984 & 0.885 & 0.26 \\
\midrule
\multirow{3}{*}{Synthetic}
 & Discriminative & 0.997 & 0.996 & 0.966 & 0.722 & 0.34 \\
 & Refined        & 0.995 & 0.995 & 0.944 & 0.695 & 0.33 \\
 & w/o margin     & 1.000 & 0.998 & 0.998 & 0.565 & 0.36 \\
\midrule
\multirow{2}{*}{CALVIN}
 & Discriminative & 0.923 & 0.901 & 0.717 & 0.579 & 0.20 \\
 & Refined        & 0.904 & 0.895 & 0.708 & 0.568 & 0.19 \\
\bottomrule
\end{tabular}%
}
\end{table}

\subsection{Naval Dataset}

\begin{figure}[!ht]
\centering
\includegraphics[width=0.88\columnwidth]{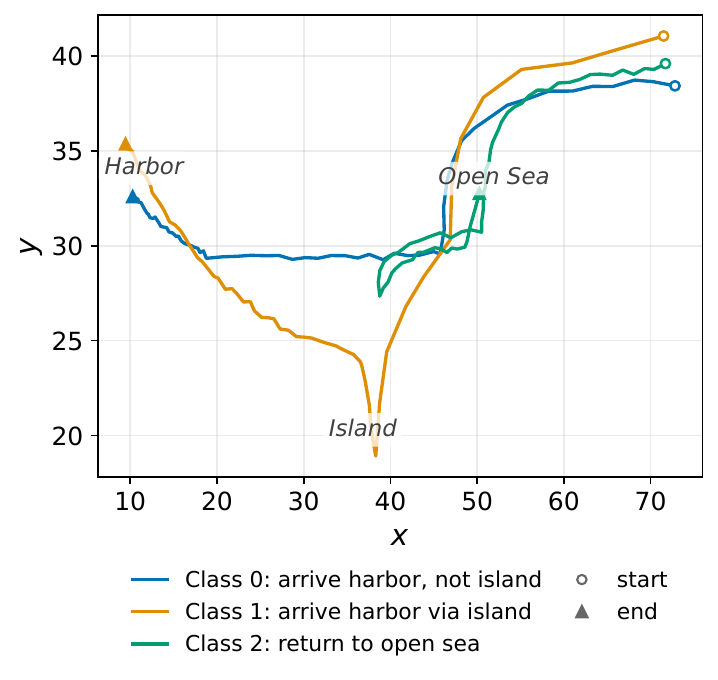}
\caption{Trajectory examples in the naval surveillance scenario. Vessels of class 0 arrive at the harbor without approaching the island, vessels of class 1 pass the island region on the way to the harbor, and vessels of class 2 turn back to the open sea. The formulas of Table~\ref{tab:formulas-naval} correspond to the regions and time windows that separate the three behaviors.}
\label{fig:dataset-naval}
\end{figure}

\begin{table}[!t]
\centering
\caption{Learned STL formulas on the naval dataset after refinement, in the original coordinate frame. Vacuous and redundant atoms are pruned using the training and validation splits, on which the pruned formulas agree with the full model by construction; on the test split they agree on all but one sample.}
\label{tab:formulas-naval}
\begin{tabular}{ll}
\toprule
Class & Formula \\
\midrule
0 & $F_{[29,31]}(x < 38.7) \wedge F_{[1,2]}(y < 44.3)$ \\
  & $\;\wedge\; F_{[1,2]}(y > 34.7) \wedge F_{[9,26]}(y > 29.2)$ \\
\midrule
1 & $G_{[5,6]}(x < 55.1) \wedge G_{[43,45]}(y > 26.6)$ \\
\midrule
2 & $G_{[41,48]}(y < 38.8) \wedge G_{[41,56]}(x > 37.1)$ \\
\bottomrule
\end{tabular}
\end{table}

The naval dataset~\cite{kong2014temporal,bombara2016decision} contains vessel trajectories of three behaviors, shown in Fig.~\ref{fig:dataset-naval}. Table~\ref{tab:main} shows that the discriminative model is accurate while some required formulas are violated, so accuracy alone overstates how well the formulas describe their classes. Refinement tightens the formulas substantially on this dataset. Among required formulas satisfied before and after
refinement, the median robustness falls from $0.17$ to $0.15$ and the
upper quartile from $0.31$ to $0.16$, so the loosest formulas are pulled toward the target of the tightening term. The cost is a moderate loss of satisfaction, and Table~\ref{tab:sweep} shows that the satisfaction weight controls how much is lost. Removing the margin reward raises both satisfaction rates without reducing accuracy, so on this dataset the reward costs satisfaction that accuracy does not reveal.

Table~\ref{tab:formulas-naval} lists the refined formulas after pruning, which keep only a few constraints per class. The class 0 formula constrains the early approach to the harbor, the class 1 formula separates the island route with two always constraints, and the class 2 formula requires the vessel to remain in the open sea region late in the horizon. Pruning changes the prediction of a single
test sample, so these few constraints carry the classification.

Since the score vanishes whenever the outputs agree with the codeword, most calibration scores are exactly zero after refinement. Pooled calibration leaves one class at coverage $0.794$, below the target, whereas class conditional calibration brings all three classes to about $0.92$ at nearly the same set size, as Fig.~\ref{fig:cp-all} shows.

\begin{figure}[!t]
\centering
\includegraphics[width=\columnwidth]{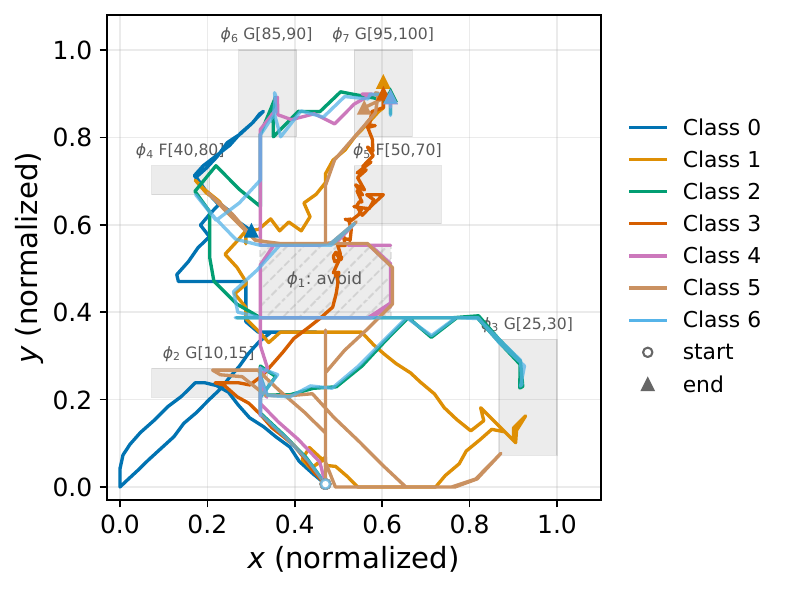}
\caption{Trajectory examples in Synthetic Dataset.}
\label{fig:dataset-synthetic}
\end{figure}

\subsection{Synthetic Dataset}

\begin{table}[!t]
\centering
\caption{Learned STL formulas on the synthetic dataset after
refinement, in the generator coordinate frame. Same pruning
convention as Table~\ref{tab:formulas-naval}.}
\label{tab:formulas-synthetic}
\begin{tabular}{ll}
\toprule
Class & Formula \\
\midrule
0 & $F_{[33,35]}(y < -2.0) \wedge G_{[13,63]}(x < 3.1)$ \\
\midrule
1 & $G_{[17,19]}(x > 3.5)$ \\
\midrule
2 & $F_{[69,70]}(y < -1.1)$ \\
\midrule
3 & $F_{[25,26]}(-3.6 < x < 2.6) \wedge G_{[75,76]}(x > 0.9)$ \\
\midrule
4 & $F_{[32,36]}(x < 1.6) \wedge G_{[62,63]}(y < 0.1)$ \\
\midrule
5 & $F_{[22,26]}(x > -4.1) \wedge G_{[38,39]}(x < 0.0)$ \\
\midrule
6 & $F_{[26,27]}(y < -3.4) \wedge F_{[23,25]}(y > -1.5)$ \\
  & $\;\wedge\; F_{[37,48]}(x > 2.2)$ \\
\bottomrule
\end{tabular}
\end{table}

The synthetic dataset~\cite{linard2022inference} contains seven classes built from combinations of spatio temporal requirements, shown in Fig.~\ref{fig:dataset-synthetic}. Because some classes satisfy all requirements of others, a formula that describes its own class cannot separate it from the classes that contain it, and discrimination must come from the required violations weighted by $\lambda_{\mathrm{neg}}$. This weight trades description against discrimination. With equal weights some classes end with no satisfied samples while accuracy stays high, and without negative pressure every required formula is satisfied while CSR collapses because the formulas hold for every
class. The intermediate weight keeps both accuracy and satisfaction high, as Table~\ref{tab:main} shows, with the remaining loss in CSR concentrated on two formulas.

On this dataset and on CALVIN, the selection criteria retain a checkpoint immediately after the start of refinement, so the refined models differ little from the discriminative ones. The pruned formulas in Table~\ref{tab:formulas-synthetic} use few constraints and keep the conditions the decoder needs rather than the full generating requirements, with the rest of the class separation carried by the required violations.

Pooled calibration covers six classes at or above the target but leaves the class with the largest scores at coverage $0.460$. Class conditional calibration raises it to $0.872$ with similar set sizes, as Fig.~\ref{fig:cp-all} shows.

\begin{figure}[t]
\centering
\includegraphics[width=\columnwidth]{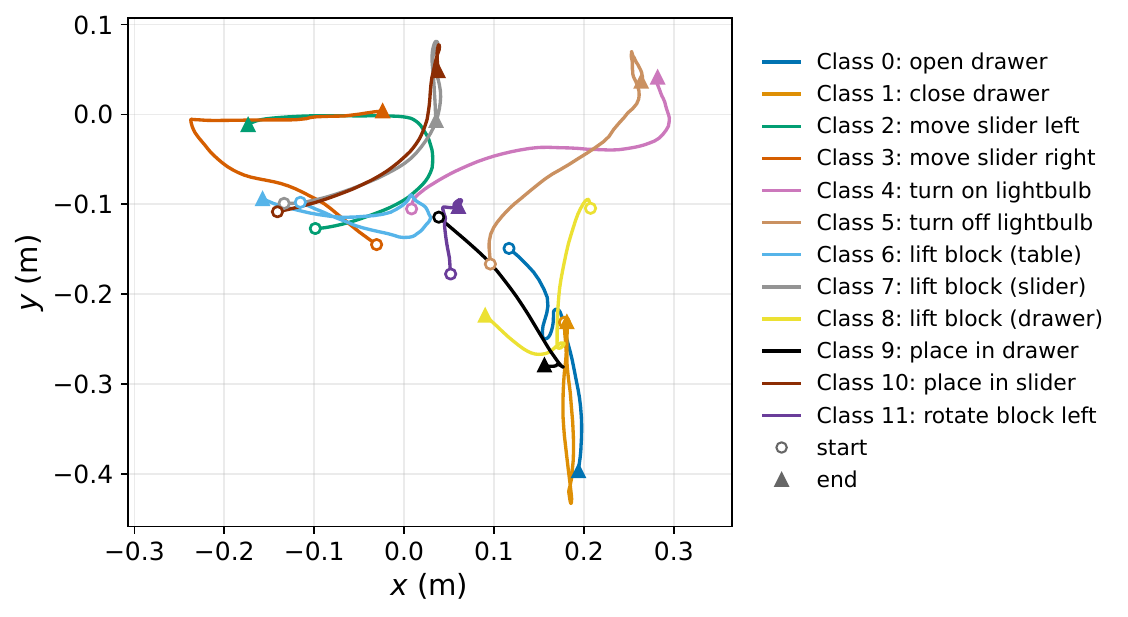}
\caption{One trajectory per task of CALVIN Dataset.}
\label{fig:dataset-calvin}
\end{figure}

\begin{table}[!t]
\centering
\caption{Learned STL formulas on the CALVIN dataset after refinement.
Same pruning convention as Table~\ref{tab:formulas-naval}. Here $x$,
$y$, $z$ are the end effector position and $g$ the gripper width, in
centimeters.}
\label{tab:formulas-calvin}

\scriptsize
\setlength{\tabcolsep}{3pt}

\resizebox{\columnwidth}{!}{%
\begin{tabular}{ll}
\toprule
Class & Formula \\
\midrule
0 & $F_{[62,63]}(y < -35.1)$ \\
\midrule
1 & $G_{[31,36]}(y < -34.0)$ \\
\midrule
2 & $F_{[38,42]}(x > -11.3 \wedge y > -2.6 \wedge g > 1.9)$ \\
  & $\;\wedge\; F_{[20,27]}(g > 4.7) \wedge G_{[7,51]}(x < 23.6 \wedge y < 3.5)$ \\
\midrule
3 & $F_{[60,63]}(x > -16.9) \wedge G_{[33,39]}(x < -5.0 \wedge y > -6.1)$ \\
  & $\;\wedge\; G_{[5,38]}(g > -0.1)$ \\
\midrule
4 & $F_{[39,52]}(y > -3.4 \wedge z < 53.4) \wedge G_{[40,44]}(x > 15.2)$ \\
\midrule
5 & $F_{[40,48]}(z > 62.6)$ \\
\midrule
6 & $F_{[31,35]}(6.3 < g < 8.1) \wedge F_{[10,11]}(x < 19.2 \wedge g < 8.1)$ \\
  & $\;\wedge\; G_{[6,7]}(z > 52.2) \wedge G_{[32,33]}(48.6 < z < 54.5)$ \\
\midrule
7 & $F_{[23,25]}(g > 6.6) \wedge F_{[42,49]}(x < 12.1 \wedge y > 6.0)$ \\
\midrule
8 & $F_{[31,41]}(y < -23.4) \wedge G_{[26,45]}(g > 3.4) \wedge G_{[19,21]}(g > 7.1)$ \\
\midrule
9 & $F_{[31,32]}(g < 7.6) \wedge G_{[27,61]}(y < -11.1) \wedge G_{[12,16]}(3.1 < g < 6.1)$ \\
\midrule
10 & $G_{[22,24]}(g < 5.8) \wedge G_{[46,47]}(y > 1.8)$ \\
\midrule
11 & $F_{[44,46]}(y < -8.7 \wedge g > 1.6) \wedge G_{[8,53]}(49.2 < z < 58.0)$ \\
\bottomrule
\end{tabular}%
}
\end{table}
\begin{table}[!t]
\centering
\caption{Network and exact STL semantics on the test split for the
selected refined models. Robustness signs agree on every code entry,
so PSR and CSR coincide under both semantics; accuracies differ
because decoding compares loss magnitudes.}
\label{tab:semantics}
\begin{tabular}{lccc}
\toprule
Dataset & Acc network & Acc exact & Sign disagreement \\
\midrule
Naval & 0.960 & 0.980 & 0/753 \\
Synthetic & 0.995 & 0.964 & 0/15337 \\
CALVIN & 0.895 & 0.871 & 0/5472 \\
\bottomrule
\end{tabular}
\end{table}

\subsection{CALVIN Dataset}

The CALVIN dataset~\cite{mees2022calvin} provides end effector trajectories of twelve manipulation tasks, shown in Fig.~\ref{fig:dataset-calvin}. It gives the clearest example of the gap between accuracy and satisfaction. Most classes satisfy their formulas, but for the rotate class the learned height constraint lies slightly beyond the range of the recorded trajectories, so no test trajectory of this class satisfies its formula, while $71\%$ of them are still classified correctly from the relative decoding losses. The pruned formulas in Table~\ref{tab:formulas-calvin} constrain the channels each task uses, the horizontal position for the drawer and slider tasks, the height for the lightbulb tasks, and the gripper width together with a position constraint for the lift and place tasks.

Class conditional calibration raises the lowest class coverage from $0.750$ to $0.870$, as Fig.~\ref{fig:cp-all} shows. With few calibration samples per class, the class thresholds are coarser than the pooled one and the realized coverage varies more, so the correction enlarges the average prediction set from $1.13$ to $1.25$.

\begin{table}[!t]
\centering
\caption{Stage 2 sweep on the naval dataset, all metrics on the test
split. The baseline row is the configuration used on the synthetic
and CALVIN datasets, and the $\eta_{\mathrm{sign}} \times 4$ row is
the reported operating point.}
\label{tab:sweep}

\setlength{\tabcolsep}{3.5pt}
\small
\begin{tabular}{lccccccc}
\toprule
Configuration & $\eta_{\mathrm{tight}}$ & $\eta_{\mathrm{sign}}$ &
$\gamma$ & Acc & PSR & CSR & $\bar{R}_{\mathrm{pos}}$ \\
\midrule
Baseline & 1.0 & 5 & 0.01 & 0.976 & 0.757 & 0.729 & 0.105 \\
$\eta_{\mathrm{tight}} \times 0.5$ & 0.5 & 5 & 0.01 & 0.976 & 0.705 & 0.661 & 0.097 \\
$\eta_{\mathrm{sign}} \times 2$ & 1.0 & 10 & 0.01 & 0.960 & 0.853 & 0.793 & 0.124 \\
$\eta_{\mathrm{sign}} \times 4$ & 1.0 & 20 & 0.01 & 0.960 & 0.873 & 0.797 & 0.128 \\
$\gamma \times 2$ & 1.0 & 5 & 0.02 & 0.976 & 0.809 & 0.777 & 0.116 \\
\bottomrule
\end{tabular}
\end{table}
\section{CONCLUSION}

This work shows that multiclass STL prediction, specification satisfaction, and uncertainty should not be treated as interchangeable notions. Multiclass decoding can remain accurate even when the robustness signs fail to realize the intended class codeword. PSR and CSR make this discrepancy explicit, while the refinement procedure exposes a controllable tradeoff between moving satisfied outputs toward the STL boundary and limiting losses in satisfaction and class separation. The same decoding geometry also provides a natural conformal score: its zero level corresponds to codeword consistency, whereas class-conditional calibration addresses the classwise coverage imbalance hidden by a pooled threshold. Across the three datasets, the experiments show that the magnitude of attainable tightening is data dependent, while the distinction between decoding accuracy and specification satisfaction persists across all three datasets. Future work should replace soft satisfaction penalties with explicit constraints and extend the framework beyond one-hot codes.

\bigskip
\noindent










\FloatBarrier
\bibliographystyle{IEEEtran}
\bibliography{reference}

@inproceedings{bombara2016decision,
  title={A decision tree approach to data classification using signal temporal logic},
  author={Bombara, Giuseppe and Vasile, Cristian-Ioan and Penedo, Francisco and Yasuoka, Hirotoshi and Belta, Calin},
  booktitle={Proceedings of the 19th International Conference on Hybrid Systems: Computation and Control},
  pages={1--10},
  year={2016}
}

@inproceedings{mohammadinejad2020interpretable,
  title={Interpretable classification of time-series data using efficient enumerative techniques},
  author={Mohammadinejad, Sara and Deshmukh, Jyotirmoy V and Puranic, Aniruddh G and Vazquez-Chanlatte, Marcell and Donz{\'e}, Alexandre},
  booktitle={Proceedings of the 23rd International Conference on Hybrid Systems: Computation and Control},
  pages={1--10},
  year={2020}
}

@article{li2024tlinet,
  title={Tlinet: Differentiable neural network temporal logic inference, 2024},
  author={Li, D and Cai, M and Vasile, CI and Tron, R},
  journal={URL https://arxiv. org/abs/2405.06670},
  year={2024}
}

@inproceedings{yan2022neuro,
  title={Neuro-symbolic models for interpretable time series classification using temporal logic description},
  author={Yan, Ruixuan and Ma, Tengfei and Fokoue, Achille and Chang, Maria and Julius, Agung},
  booktitle={2022 IEEE International Conference on Data Mining (ICDM)},
  pages={618--627},
  year={2022},
  organization={IEEE}
}

@inproceedings{li2024multi,
  title={Multi-class temporal logic neural networks},
  author={Li, Danyang and Tron, Roberto},
  booktitle={2024 American Control Conference (ACC)},
  pages={5155--5162},
  year={2024},
  organization={IEEE}
}

@book{vovk2005algorithmic,
  title={Algorithmic learning in a random world},
  author={Vovk, Vladimir and Gammerman, Alexander and Shafer, Glenn},
  year={2005},
  publisher={Springer}
}

@article{romano2020classification,
  title={Classification with valid and adaptive coverage},
  author={Romano, Yaniv and Sesia, Matteo and Candes, Emmanuel},
  journal={Advances in neural information processing systems},
  volume={33},
  pages={3581--3591},
  year={2020}
}

@article{huang2023conformal,
  title={Conformal prediction for deep classifier via label ranking},
  author={Huang, Jianguo and Xi, Huajun and Zhang, Linjun and Yao, Huaxiu and Qiu, Yue and Wei, Hongxin},
  journal={arXiv preprint arXiv:2310.06430},
  year={2023}
}

@article{gibbs2025conformal,
  title={Conformal prediction with conditional guarantees},
  author={Gibbs, Isaac and Cherian, John J and Cand{\`e}s, Emmanuel J},
  journal={Journal of the Royal Statistical Society Series B: Statistical Methodology},
  volume={87},
  number={4},
  pages={1100--1126},
  year={2025},
  publisher={Oxford University Press UK}
}

@article{mees2022calvin,
  title={Calvin: A benchmark for language-conditioned policy learning for long-horizon robot manipulation tasks},
  author={Mees, Oier and Hermann, Lukas and Rosete-Beas, Erick and Burgard, Wolfram},
  journal={IEEE Robotics and Automation Letters},
  volume={7},
  number={3},
  pages={7327--7334},
  year={2022},
  publisher={IEEE}
}

@inproceedings{aasi2023time,
  title={Time-incremental learning of temporal logic classifiers using decision trees},
  author={Aasi, Erfan and Cai, Mingyu and Vasile, Cristian Ioan and Belta, Calin},
  booktitle={Learning for Dynamics and Control Conference},
  pages={547--559},
  year={2023},
  organization={PMLR}
}

@article{liu2024interpretable,
  title={Interpretable imitation learning via generative adversarial STL inference and control},
  author={Liu, Wenliang and Li, Danyang and Aasi, Erfan and Rus, Daniela and Tron, Roberto and Belta, Calin},
  journal={arXiv preprint arXiv:2402.10310},
  year={2024}
}

@article{ferfoglia2025towards,
  title={Towards Interpretable Concept Learning over Time Series via Temporal Logic Semantics},
  author={Ferfoglia, Irene and Silvetti, Simone and Saveri, Gaia and Nenzi, Laura and Bortolussi, Luca},
  journal={arXiv preprint arXiv:2508.03269},
  year={2025}
}

@article{liu2024timex++,
  title={Timex++: Learning time-series explanations with information bottleneck},
  author={Liu, Zichuan and Wang, Tianchun and Shi, Jimeng and Zheng, Xu and Chen, Zhuomin and Song, Lei and Dong, Wenqian and Obeysekera, Jayantha and Shirani, Farhad and Luo, Dongsheng},
  journal={arXiv preprint arXiv:2405.09308},
  year={2024}
}

@article{soroka2024learning,
  title={Learning temporal logic predicates from data with statistical guarantees},
  author={Soroka, Emi and Sinha, Rohan and Lall, Sanjay},
  journal={arXiv preprint arXiv:2406.10449},
  year={2024}
}

@article{jha2019telex,
  title={TeLEx: learning signal temporal logic from positive examples using tightness metric},
  author={Jha, Susmit and Tiwari, Ashish and Seshia, Sanjit A and Sahai, Tuhin and Shankar, Natarajan},
  journal={Formal Methods in System Design},
  volume={54},
  number={3},
  pages={364--387},
  year={2019},
  publisher={Springer}
}

@article{li2025conformal,
  title={Conformal prediction for signal temporal logic inference},
  author={Li, Danyang and Wang, Yixuan and Cleaveland, Matthew and Cai, Mingyu and Tron, Roberto},
  journal={arXiv preprint arXiv:2509.25473},
  year={2025}
}

@article{bairaktari2025kandinsky,
  title={Kandinsky conformal prediction: Beyond class-and covariate-conditional coverage},
  author={Bairaktari, Konstantina and Wu, Jiayun and Wu, Zhiwei Steven},
  journal={arXiv preprint arXiv:2502.17264},
  year={2025}
}

@article{shi2024conformal,
  title={Conformal prediction for class-wise coverage via augmented label rank calibration},
  author={Shi, Yuanjie and Ghosh, Subhankar and Belkhouja, Taha and Doppa, Janardhan R and Yan, Yan},
  journal={Advances in Neural Information Processing Systems},
  volume={37},
  pages={132133--132178},
  year={2024}
}

@inproceedings{gao2025bridging,
  title={Bridging fairness and efficiency in conformal inference: A surrogate-assisted group-clustered approach},
  author={Gao, Chenyin and Gilbert, Peter B and Han, Larry},
  booktitle={Forty-second International Conference on Machine Learning},
  year={2025}
}

@article{dietterich1994solving,
  title={Solving multiclass learning problems via error-correcting output codes},
  author={Dietterich, Thomas G and Bakiri, Ghulum},
  journal={Journal of artificial intelligence research},
  volume={2},
  pages={263--286},
  year={1994}
}

@article{allwein2000reducing,
  title={Reducing multiclass to binary: A unifying approach for margin classifiers},
  author={Allwein, Erin L and Schapire, Robert E and Singer, Yoram},
  journal={Journal of machine learning research},
  volume={1},
  number={Dec},
  pages={113--141},
  year={2000}
}

@inproceedings{kong2014temporal,
  title={Temporal logic inference for classification and prediction from data},
  author={Kong, Zhaodan and Jones, Austin and Medina Ayala, Ana and Aydin Gol, Ebru and Belta, Calin},
  booktitle={Proceedings of the 17th international conference on Hybrid systems: computation and control},
  pages={273--282},
  year={2014}
}

@inproceedings{linard2022inference,
  title={Inference of multi-class stl specifications for multi-label human-robot encounters},
  author={Linard, Alexis and Torre, Ilaria and Leite, Iolanda and Tumova, Jana},
  booktitle={2022 IEEE/RSJ International Conference on Intelligent Robots and Systems (IROS)},
  pages={1305--1311},
  year={2022},
  organization={IEEE}
}

@inproceedings{maler2004monitoring,
  title={Monitoring temporal properties of continuous signals},
  author={Maler, Oded and Nickovic, Dejan},
  booktitle={International symposium on formal techniques in real-time and fault-tolerant systems},
  pages={152--166},
  year={2004},
  organization={Springer}
}

@inproceedings{donze2010robust,
  title={Robust satisfaction of temporal logic over real-valued signals},
  author={Donz{\'e}, Alexandre and Maler, Oded},
  booktitle={International conference on formal modeling and analysis of timed systems},
  pages={92--106},
  year={2010},
  organization={Springer}
}

@ARTICLE{7084172,
  author={Jin, Xiaoqing and Donzé, Alexandre and Deshmukh, Jyotirmoy V. and Seshia, Sanjit A.},
  journal={IEEE Transactions on Computer-Aided Design of Integrated Circuits and Systems}, 
  title={Mining Requirements From Closed-Loop Control Models}, 
  year={2015},
  volume={34},
  number={11},
  pages={1704-1717},
  doi={10.1109/TCAD.2015.2421907}}

@article{leung2023backpropagation,
  title={Backpropagation through signal temporal logic specifications: Infusing logical structure into gradient-based methods},
  author={Leung, Karen and Ar{\'e}chiga, Nikos and Pavone, Marco},
  journal={The International Journal of Robotics Research},
  volume={42},
  number={6},
  pages={356--370},
  year={2023},
  publisher={SAGE Publications Sage UK: London, England}
}

@ARTICLE{9662049,
  author={Baharisangari, Nasim and Hirota, Kazuma and Yan, Ruixuan and Julius, Agung and Xu, Zhe},
  journal={IEEE Control Systems Letters}, 
  title={Weighted Graph-Based Signal Temporal Logic Inference Using Neural Networks}, 
  year={2022},
  volume={6},
  number={},
  pages={2096-2101},
  doi={10.1109/LCSYS.2021.3138059}}

@INPROCEEDINGS{8815145,
  author={Xu, Zhe and Ornik, Melkior and Julius, A. Agung and Topcu, Ufuk},
  booktitle={2019 American Control Conference (ACC)}, 
  title={Information-Guided Temporal Logic Inference with Prior Knowledge}, 
  year={2019},
  volume={},
  number={},
  pages={1891-1897},
  doi={10.23919/ACC.2019.8815145}}

@inproceedings{baharisangari2021uncertainty,
  title={Uncertainty-aware signal temporal logic inference},
  author={Baharisangari, Nasim and Gaglione, Jean-Rapha{\"e}l and Neider, Daniel and Topcu, Ufuk and Xu, Zhe},
  booktitle={International Workshop on Numerical Software Verification},
  pages={61--85},
  year={2021},
  organization={Springer}
}

\clearpage

\end{document}